%% file: letter.tex
\documentclass[11pt,a4paper]{article} 
\input{letter_setting}

\graphicspath{{./figs/}}
\usepackage{booktabs} 
\usepackage{times}
\usepackage{graphicx} 
\usepackage{subfigure}
\usepackage{amsfonts}
\usepackage{multirow}
\usepackage{array}
\usepackage{amsmath}
\usepackage{algorithm}
\usepackage{algorithmic}
\usepackage{color}
\begin{document}

\LetterHeader{1}{1}{February 1, 2018}  
\LetterTitle{TC-CCPS Newsletter}       
\clearpage

\SectionTitle{Technical Articles}
\input{item1}

\end{document}

%% file: letter_setting.tex
\usepackage{caption}         %
\usepackage{charter}         
\usepackage{graphicx}        
\usepackage{amssymb,amsmath} 
\usepackage{multicol}        
\usepackage{url}             
\usepackage{enumitem}        
\usepackage{marvosym}        
\usepackage{wrapfig}         
\usepackage[T1]{fontenc}     
\usepackage{mathptmx}        
\usepackage{datetime}        
\usepackage{fancyhdr}        
\newdateformat{mydate}{\monthname[\THEMONTH] \THEYEAR}          
\usepackage[pdfpagemode=FullScreen, colorlinks=false]{hyperref} 

\newcommand{\HorRule}[1]{\noindent\rule{\linewidth}{#1}}

\newcommand{\LetterTitle}[1]
{
    \begin{center}
    \usefont{T1}{ptm}{b}{n}  
    \Huge #1
    \end{center}	
    \par \normalsize \normalfont
    \HorRule{3pt}
}

\newcommand{\LetterHeader}[3]
{
    \usefont{T1}{ptm}{n}{n}  
    \hfill \textsc{Volume #1, Issue #2, #3}  
    \par \normalsize \normalfont
}

\newcommand{\SectionTitle}[1]
{
    \HorRule{1pt}
    \usefont{T1}{ptm}{b}{n}               
    \vspace{3pt}\Large #1\vspace{1pt}    
    \par \normalsize \normalfont
    \HorRule{1pt}
}

\newcommand{\NewsTitle}[1]
{
    \begin{center}
    \usefont{T1}{ptm}{b}{n}               
    \vspace{14pt}\Large #1\vspace{1pt}    
    \par \normalsize \normalfont
    \end{center}
}

\newcommand{\NewsAuthor}[1]
{
    \begin{center}
    \usefont{T1}{ptm}{n}{n}               
    \textsc{#1} \vspace{4pt} 
    \par \normalfont
    \end{center}
}

%% file: item1.tex
\NewsTitle{On the Quantization of Cellular Neural Networks for Cyber-Physical Systems}
\NewsAuthor{Xiaowei Xu, University of Notre Dame}

\section{Introduction and Motivation}

Cyber-Physical Systems (CPSs) have been pervasive including smart grid, autonomous automobile systems, medical monitoring, process control systems, robotics systems, and automatic pilot avionics \cite{cpsSurvey}\cite{xu2018dac}\cite{xu2018accelerating}\cite{xu2018scaling}\cite{xu2018quantization}.
As usually implemented on embedded devices, CPS is typically constrained by computation capacity and energy consumption.
In some CPS applications such as telemedicine \cite{xu2017segmentation}\cite{liu2018efficient} and advanced driving assistance system (ADAS) \cite{feiden2002obstacle}, data processing on the embedded devices is preferred due to security/safety and real-time requirement\cite{xu2018mda}.
Therefore, high efficiency is highly desirable for such CPS applications.

A very powerful tool for telemedicine and ADAS is cellular neural network (CeNN), which can achieve very high accuracy through proper training.
It should be noted that CeNNs are popular in image processing areas such as classification \cite{dohler2008cellular}, segmentation \cite{duraisamy2014cellular}, while convolutional Neural Networks (CNNs) are most powerful in classification related tasks.
However, due to the complex nature of segmentation and other image process tasks and the associated real-time requirements in many applications, hardware implementations of CeNNs have remained an active research topic in the literature.

The structure of CeNNs makes them a natural fit for analog implementations.
Many studies exist along this direction  \cite{harrer1994current}\cite{rodriguez2004ace16k}\cite{manganaro2012cellular}\cite{carey2013mixed}.
The advantages of analog implementations include high performance with an extremely fast convergence rate and the convenience of integrating them into image sensors for direct processing of captured data.
However, these analog implementations suffer from Input/Output (I/O) and data precision problems.
First, they require that each input corresponds to a unique neuron cell, resulting in too many I/O ports.
For example, recent implementation \cite{carey2013mixed} can only support 256$\times $256 pixels at its most, which is far from the requirement of mainstream images, e,g., 1920$\times$1080 pixels.
Second, analog circuits are prone to noise, which limit the output data precision to 7 bits or below \cite{yildiz2015architecture}.
As a result, analog implementation cannot even process regular 8-bit gray images.

In view of the above issues, digital implementations of CeNNs have been proposed, where data is quantized with approximation.
Tens to hundreds of iterations are needed in the discretized process and as a result, the computational complexity of digital CeNNs is very high.
For example, to process an image of 1920x1080 pixels requires 4-8 Giga operations (for 3$\times$3 templates and 50-100 iterations), which needs to be done in a timely manner for real-time processing.

To tackle the computation challenge, CeNN accelerations on digital platforms such as ASICs \cite{lee201124}\cite{manatunga2015sp}, GPUs \cite{potluri2011cnn} and FPGAs \cite{chen2006image}\cite{porter2007reconfigurable} \cite{martinez2013efficient}\cite{yildiz2015architecture}\cite{yildiz2016way11111}\cite{xu2018efficient}\cite{muller2016improved} have been explored, with FPGA among the most popular choices due to its high flexibility and low time-to-market.
The work \cite{chen2006image} presented a baseline design with several applications, while the study \cite{porter2007reconfigurable} took advantage of reconfigurable computing for CeNNs.
Recently, the CeNN implementation for binary images was demonstrated \cite{muller2016improved}.
Expandable and pipelined implementations were achieved on multiple FPGAs \cite{martinez2013efficient}.
Taking advantage of the structure in \cite{martinez2013efficient}, the work \cite{yildiz2015architecture} implemented a high throughput real-time video streaming system, which is further improved to be a complete system for video processing \cite{yildiz2016way11111}.
All the three works share the same architecture for CeNN computation.
Due to the large number of multiplications needed in CeNNs, the limited number of embedded multipliers in an FPGA become the bottleneck for further improvement.
For example, in work \cite{martinez2013efficient} 95\%-100\% of the embedded multipliers are used.
On the other hand, it is interesting to note that the utilization rates of LEs and registers are only 5\% and 2\%, respectively, which is natural to expect as not many logic operations are needed.
However, in a mainstream FPGA, LEs and registers count for significantly larger portion of the total programmable resources than embedded multipliers.
For example, LEs and registers occupy 95.4\% of the core area while embedded multipliers only 4.6\% for a EP3LS340 FPGA \cite{wong2011comparing}.
Such an unbalanced resource utilization apparently cannot attain the best possible speed of the CeNN being implemented, and an improved strategy is strongly desired.

A naive approach for potential improvement is to use LEs and registers to implement additional multipliers.
This technique, although straightforward, is very inefficient due to the high cost.
For example, it takes 676 LEs and 486 shift registers to implement an 18-bit multiplier. For an XC4LX25 FPGA, all the LEs and registers can only contribute 42\% additional multipliers.
Apparently, such an approach would not lead to significant improvement and we aim to address the problem through an alternative approach, i.e., by completely eliminating the need of multipliers.
From basic Boolean algebra, we know that the multiplication of any number with powers of two can simply be done with logic shift, which only requires a small number of LEs and registers to achieve.
Inspired by this, we can quantize the values in CeNN templates to powers of two, so that we can make full use of the abundant LEs and registers in FPGAs.
An extra benefit from this approach is that LEs and registers are much more flexible for placement
and routing, leading to higher clock frequencies.
While this can lead to significantly higher resource utilization rate and reduced computational complexity, many interesting questions still remain.
For example, how would such quantizations affect the final CeNN accuracy?
What is the impact of different quantization strategies?
Note that quantization to powers of two has been explored in the context of CNNs \cite{zhou2017Incremental}, but as detailed in Section \ref{motivation}, the difference in computation structures between CeNNs and CNNs warrants a separate investigation for CeNNs.
And indeed, our findings show that the answers to these questions are different for the two.

In this paper we present CeNN quantization for high-efficient processing for CPS applications, particularly telemedicine and ADAS applications.
We systematically put forward powers-of-two based incremental quantization of CeNNs for efficient hardware implementation.
The incremental quantization contains iterative procedures including parameter partition, parameter quantization, and re-training.
We propose five different strategies including random strategy, pruning inspired strategy, weighted pruning inspired strategy, nearest neighbor strategy, and weighted nearest neighbor strategy.
Experimental results show that our approach can achieve a speedup up to 7.8x with no performance loss compared with the state-of-the-art FPGA solutions for CeNNs.

The remainder of the paper is organized as follows.
Section \ref{secBackground} introduces backgrounds and motivation of the paper.
The proposed framework for CeNN and the optimized hardware implementation are presented in Section
\ref{secMethod}.
Experiments and discussion are provided in Section \ref{secExperiment} and concluding remarks are given in Section \ref{secConclusion}.

\section{Preliminaries}\label{secBackground}
\subsection{Cellular Neural Networks}
Different from the prevalent CNNs which are superior for classification tasks, the CeNN model is inspired by the functionality of visual neurons.
In a CeNN, a mass of neuron cells are connected with neighbouring ones, and only adjacent cells can interact directly with each other.
This is a significant advantage for hardware implementation, resulting in much less routing complexity and area overhead.
CeNNs are superior at image processing tasks that involve sensory functions, such as noise cancellation, edge detection, path planning, segmentation, etc.
For the widely used 2D CeNN with space-invariant templates, the dynamics of each cell state with an M$\times$N rectangular cell array \cite{chua2002cellular} are as follows:

\vspace{-1pt}
\begin{equation}\label{formula:CeNN1}
\begin{split}
\dot{x}_{i,j}(t)=-x_{i,j}(t)+\sum_{k,l=-N}^{N} (A_{k,l}(t)y_{i+k,j+l}(t)+\\
B_{k,l}(t)u_{i+k,j+l}(t))+I(t),
\end{split}
\end{equation}
{
\begin{equation}\label{formula:CeNN2}
y_{i,j}(t)=f(x_{i,j}(t))=0.5\times (|x_{i,j}(t)+1|-|x_{i,j}(t)-1|),
\end{equation}
\vspace{-1pt}
}

\noindent where $1\leq i\leq M$, $1\leq j\leq N$, $A_{k,l}(t)$ is the feedback coefficient template, $B_{k,l}(t)$ is the feedforward coefficient template, $I(t)$ is the bias, and $x_{i,j}(t)$, $y_{i+k,j+l}(t)$ and $u_{i+k,j+l}(t)$ are the state, output and input of the cell, respectively.
Note that $A_{k,l}(t)$, $B_{k,l}(t)$ and $I(t)$ are time-variant templates, and $t$ can be removed when time-invariant templates are used.
For efficient implementation on a digital platform (e.g., CPU, GPU, FPGA), discrete approximation of CeNN is obtained by applying forward Euler approximation as shown in Equations \ref{formula:CeNND1}, \ref{formula:CeNND2} and \ref{formula:CeNND3}.

{
\begin{equation}\label{formula:CeNND1}
{x}_{i,j}(t)\cong {(x_{i,j}(n+1)-x_{i,j}(n))}/{\Delta t}.
\end{equation}
}
{
\begin{equation}\label{formula:CeNND2}
\begin{split}
{x}_{i,j}(n+1)=x_{i,j}(n)+\Delta t(-{x}_{i,j}(n)+I(n)+\sum_{k,l=-N}^{N} (\\
A_{k,l}(n)y_{i+k,j+l}(n)+B_{k,l}(n)u_{i+k,j+l}(n))).
\end{split}
\end{equation}
}
{
\begin{equation}\label{formula:CeNND3}
y_{i,j}(n)=f(x_{i,j}(n))=0.5\times (|x_{i,j}(n)+1|-|x_{i,j}(n)-1|).
\end{equation}
}

Delayed CeNN is a special type of CeNN described by adding $\sum_{k,l=-N}^{N} (D_{i,j}(n)g(x_{k,l}(n),y_{k,l}(n),u_{k,l}(n))$ to Equation \ref{formula:CeNND2}, where $g$ is usually a piece-wise constant function.
Please refer to \cite{chua2002cellular} for details.
For the mainstream image size with 1920$\times$1080 pixels, the total complexity is 1920$\times$1080$\times$39$\times$100=8.1$\times 10^9$ operations with 100 iterations (19 multiplications and 20 additions in each iteration).
This warrants exploration of hardware approaches to speedup CeNN computations.

\subsection{Template Learning Algorithm and PSO Algorithm}

Template learning is a widely applied method to find satisfactory templates for CeNN-based applications, in which Genetic Algorithm (GA) and Particle Swarm Optimization (PSO) are two representatives.
PSO is adopted in this paper, while GA and other template learning methods are also compatible with the framework proposed here.

PSO finds solutions (i.e., determining A, B and I templates) in a heuristic way by searching the solution space with multiple particles (swarm of potential solutions).
In each iteration, PSO performs position update and object function calculation. 
Inspired by the social behavior of animals, the position update of each particle is affected by its past best position and the position of the current global best position as depicted by Equation \ref{formula:PSO},

\begin{equation}\label{formula:PSO}
\vspace{-2pt}
\begin{split}
p_{i,d}(n+1)=p_{i,d}(n)+\{w\times v_{i,d}(n)+c_1 r_1 \\
\times (pb_{i,d}-p_{i,d}(n))+c_2 r_2\times (gb_d-p_{i,d}(n))\}.
\end{split}
\end{equation}

\noindent where $1\leq i\leq N$, $1\leq d\leq D$, $N$ is the size of particles, $D$ is the dimension of each particle, $c_1$ and $c_2$ are the acceleration coefficients, and $r_1$ and $r_2$ are random numbers with uniform distribution.
$p_{i}(n+1)$ and $p_{i}(n)$ are the positions of the $i$th particle in iteration $n$ and $n+1$, respectively.
$pb_{n}$ is the best position that the $i$th particle ever searches, and $gb$ is the current best position among all particles.
Inertia weight $w$ controls the balance of the search algorithm between exploration and exploitation.
A bound of [$min_d$, $max_d$] is introduced for $p_{i,d}$ to limit the solution space.
The object function for particles taking positions as input is designed according to applications.
In CeNN training, PSO will search the space constructed with A, B and I templates, and the templates with the best object function value are obtained as the learned templates.


\begin{figure}
\begin{center}
\centerline{\includegraphics[width=0.4\columnwidth]{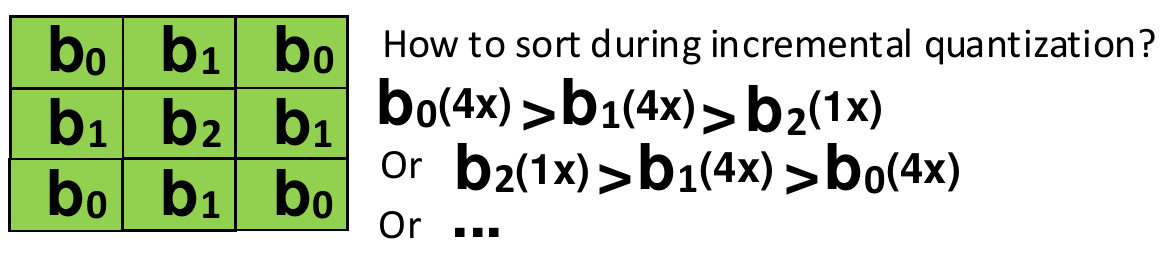}}
\vspace{-1pt}
\caption{CeNN template for binary image noise cancellation application.}
\label{fig:motivation}
\end{center}
\vspace{-1pt}
\end{figure}


{\color{black}{
\subsection{Motivation}\label{motivation}

While hardware oriented memory/computation compression and optimization of CNNs have been extensively studied recently \cite{courbariaux2016binarized}\cite{hubara2016binarized}
\cite{wu2016quantized}\cite{rastegari2016xnor}\cite{han2016deep}\cite{zhou2017Incremental}, little has been explored for CeNNs where memory consumption is not a problem and the focus is only on computational complexity.

The main difference between CeNNs and CNNs is that in CeNNs the parameters are coupled.
The weight values in a CNN tend to be all unique.
However, in CeNNs some parameters share the same values. For example, in Figure \ref{fig:motivation}, a CeNN template (template B) for binary image noise cancellation \cite{li2011edge} is shown.
Only three different values exist for the nine parameters.
As such, in \cite{zhou2017Incremental} the weights of CNNs are incrementally quantized in an order simply based on their magnitudes (pruning-inspired strategy). The same strategy may not work well for CeNNs, as a parameter with small magnitude may repeat multiple times thus playing a more important role than a parameter with a large magnitude but appearing only once. Furthermore, the training process of CNNs is mathematically optimal, while that of CeNNs is heuristic. This will also influence the performance of quantization strategies. Finally, the sparsity and repetition existing in CeNN templates provide some additional opportunity for further improvement when implemented in hardware.

\begin{figure}
\begin{center}
\centerline{\includegraphics[width=0.5\columnwidth]{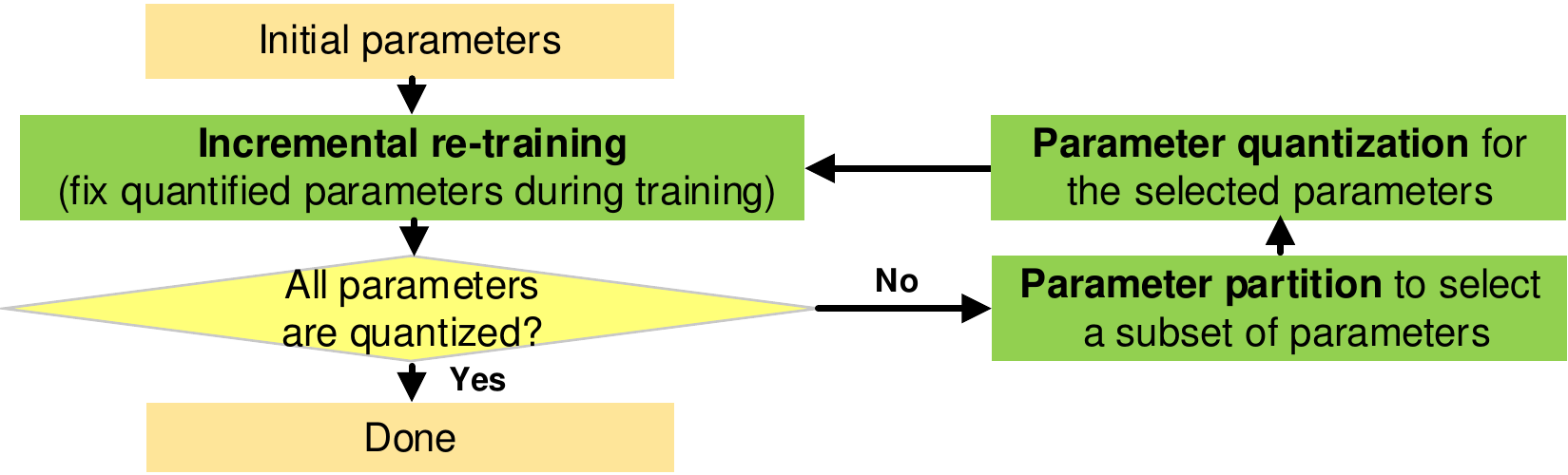}}
\vspace{-11pt}
\caption{The flowchart of incremental quantization.}
\label{fig:incrementalQuantization}
\end{center}
\vspace{-15pt}
\end{figure}

\begin{figure}
\begin{center}
\centerline{\includegraphics[width=0.6\columnwidth]{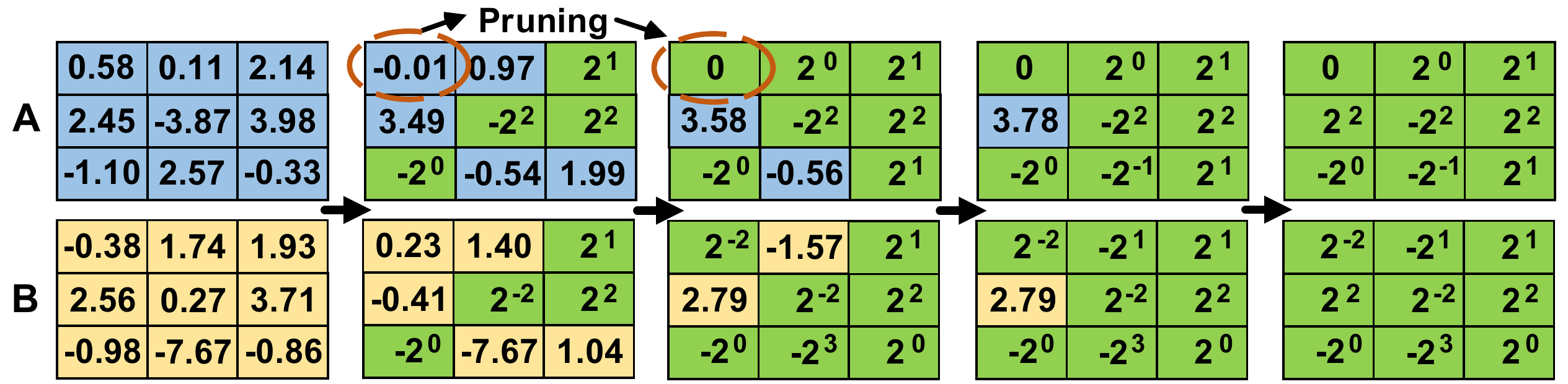}}
\vspace{-1pt}
\caption{An example of the proposed quantization framework.
In each iteration, parameter partition, parameter quantization and incremental re-training are performed sequentially.
Green cells represent quantized parameters.}
\label{fig:incrementalQuantizationIllustation}
\end{center}
\vspace{-18pt}
\end{figure}	
\section{CeNN Quantization and Hardware Implementation}\label{secMethod}
In this section, we present the CeNN quantization framework followed by the details of the hardware implementation.
\subsection{Incremental Quantization}
The proposed incremental quantization framework is an iterative process as shown in Figure \ref{fig:incrementalQuantization}.
Each iteration completes three tasks: parameter partition, parameter quantization, and incremental re-training.
We assume that as a starting point, we have all parameters in the original templates before quantization well trained.
An illustrative example of the process is shown in Figure \ref{fig:incrementalQuantizationIllustation} to facilitate understanding.


\subsubsection{Parameter Partition}
This task selects a subset of parameters not yet quantized (un-quantized parameters) to perform quantization.
Two knobs exist in this task: parameter priority and batch size.

{\color{black}{For the first knob, the pruning-inspired (PI) strategy has been well explored in quantization of CNNs \cite{zhou2017Incremental}, based on the consideration that weights with larger magnitudes contribute more to the result and thus should be quantized first.
However, the parameters in CeNNs have some unique characteristics which have been discussed in Section \ref{motivation}.
In order to tackle the problem, we propose a nearest neighbor (NN) strategy and a weighting method for the first knob.}}
The combined weighted nearest neighbor algorithm takes the number that a parameter appears in the template, defined as its repetition quantity (rq) as the reciprocal of the weight, and uses the difference between the parameter and its nearest power-of-two as distance to perform a weighted NN algorithm (WNN).
The detail explanation of WNN algorithm is shown in Algorithm \ref{alg:weightNN}.
Other combinations such as weighted pruning-inspired (WPI) strategy adopt the same weighting method but with PI to form WPI.
A total of five strategies PI, WPI, NN (WNN with all weights set to 1), WNN and a random strategy (RAN) are compared in the experimental section.

For the second knob, batch size is the number of parameters selected in each iteration, which will affect re-training speed and quality.
We propose to use two batch sizes, constant and log-scale.
The former selects the same number of parameters in each iteration, while the latter picks a fixed percentage from the remaining un-quantized parameters, rounded to the nearest integer.
Compared with constant batch size, log-scale batch size quantizes more parameters in the first several iterations and fewer towards the end.



\begin{algorithm}
   \caption{Weighted nearest neighbor strategy}
   \label{alg:weightNN}
\begin{algorithmic}
  \STATE {\bfseries Input:} un-quantized parameters $uq_i$, repeat quantity, $rq_i$, selected quantity, $N$, $1\leq i\leq n$, $n$, the number of un-quantized parameters
   \STATE {\bfseries Output:} the most important $N$ parameters
   \STATE $neighbor=log_2$ $|(uq)|$; // get the power of the absolute value of the un-quantized parameters
   \FOR{$i=1$ {\bfseries to} $n$}
   \STATE $md=(2^{floor(neighbor(i))}+2^{floor(neighbor(i)+1)})/2$;
   \IF{$md > |(uq(i))|$}
   \STATE $nnDist(i)=|(uq(i))|-2^{floor(neighbor(i))}$;
   \ELSE
   \STATE $nnDist(i)=2^{floor(neighbor(i))+1}-|(uq(i))|$;
   \ENDIF
   \ENDFOR
   \STATE $wnnDist=nnDist/ rq$;
   \STATE sort $wnnDist$ in ascending order;
   \STATE output the first $N$ parameters;
\end{algorithmic}
\end{algorithm}

\vspace{-4pt}
\subsubsection{Parameter Quantization}
Before parameter quantization, the bit width should be defined first according to applications.
Note that there are millions of parameters for CNN, and short bit width is always appreciated considering memory and computational consumption.
However, CeNN usually has tens to hundreds of parameters (time-variant templates have more parameters than time-invariant templates), and bit width has no significant impact on memory consumption.
In addition, with power-of-two conversion multiplications can be done with logic shifts, and bit width will also have little impact on computation complexity.
The only impact it will have is on the resource utilization of multipliers.

Suppose the quantization set is designed as depicted in Equation \ref{formula:quantization}, where $k$ and $m$ indicate the range of quantization.
The corresponding bit width $bw$ is calculated as shown in Equation \ref{formula:quantization2}, where the extra one bit is the sign bit.

\vspace{-1pt}
\begin{equation}\label{formula:quantization}
qs=\{\pm (2^{k},.,2^{p},.,2^{m}),0\},\;\;k\leq p\leq m,\;\;p,k,m\in \mathbb{Z}.
\vspace{-1pt}
\end{equation}


{
\vspace{-8pt}
\begin{equation}\label{formula:quantization2}
bw=Ceiling[log_2(2\times (m-k+1)+1)]+1.
\end{equation}
\vspace{-1pt}
}

With the quantization set, a parameter $uq(i)$ is quantized as shown in Equation \ref{formula:quantization3}.
When the absolute value of a parameter is smaller than $2^{-k-1}$, it will become zero after quantization and get pruned.
Lower bit width can prune more parameters, at the cost of accuracy loss.

{
\vspace{-3pt}
\begin{equation}\label{formula:quantization3}
uq(i)=\begin{cases}
2^p & \text{ if } 3\times2^{p-2}\leq |uq(i)|< 3\times2^{p-1}; \\
    &                    \;\;\;\;\;\;\;\;\;\;\;\;k\leq p\leq m;\\
2^m & \text{ if } |uq(i)|\geq 2^m; \\
0   & \text{ if } |uq(i)| < 2^{-k-1}.\\
\end{cases}
\end{equation}
\vspace{-3pt}
}

	
\subsubsection{Incremental Re-training Algorithm}
Usually, re-training algorithm is an optimal problem as shown in Equation \ref{formula:training}, where $P$ is the set of all the parameters.
In incremental re-training algorithm, the optimal problem is revised as shown in Equation \ref{formula:training2}, where $U$ and $Q$ are the sets of un-quantized and quantized parameters, respectively.
$a_i$ and $b_i$ are the lower and upper bounds for both $P_i$ and $U_i$, respectively.
Note that $P=Q\cup U$, and $U\cap Q=$$\emptyset$.
In each iteration, a subset of $U$ will be quantized and added to $Q$.

{\vspace{-12pt}
\begin{equation}\label{formula:training}
f = min\;obj(P),\;s.t.\;P_i\in [a_i, b_i], 0\leq i\leq |P|.
\end{equation}
\vspace{-1pt}
}

{
\vspace{-14pt}
\begin{equation}\label{formula:training2}
f= min\; obj(U,Q),s.t.U_i\in [a_i, b_i], 0\leq i\leq |U|.
\end{equation}
\vspace{-14pt}
}

$Q$ will be fixed during the re-training process and only $U$ is used for space searching.
After multiple iterations, all the required parameters are quantized.
It should be noted that the bias $I(n)$ in Equation \ref{formula:CeNND2} for CeNN is not required to be quantized as it is not involved in multiplication.
Therefore, another re-training iteration is required for the optimal bias when all the required parameters are quantized.

\subsection{Efficient Hardware Implementations}


We base our work on the state-of-the-art FPGA CeNN implementations \cite{martinez2013efficient}\cite{yildiz2015architecture}\cite{yildiz2016way11111}, which is expandable, highly parallel and pipelined.
The basic element of the architecture is the stage module which handles all the processes in one iteration corresponding to Equation \ref{formula:CeNND2} for $1\leq i\leq M$, $1\leq j\leq N$.
Multiple stages are connected sequentially for multiple iterations to form a layer, which processes the input in a pipelined manner.
Furthermore, multiple layers can be connected sequentially for more complex processing or be distributed in parallel for a higher throughput.
Note that First In First Out (FIFO) are used between adjacent stages to store the temporary results of each stage (or each iteration), and they are configured as single-input multiple-output memories.
Please refer to FPGA implementations in \cite{martinez2013efficient}\cite{yildiz2015architecture} for more details.

\begin{figure}
\begin{center}
\centerline{\includegraphics[width=0.65\columnwidth]{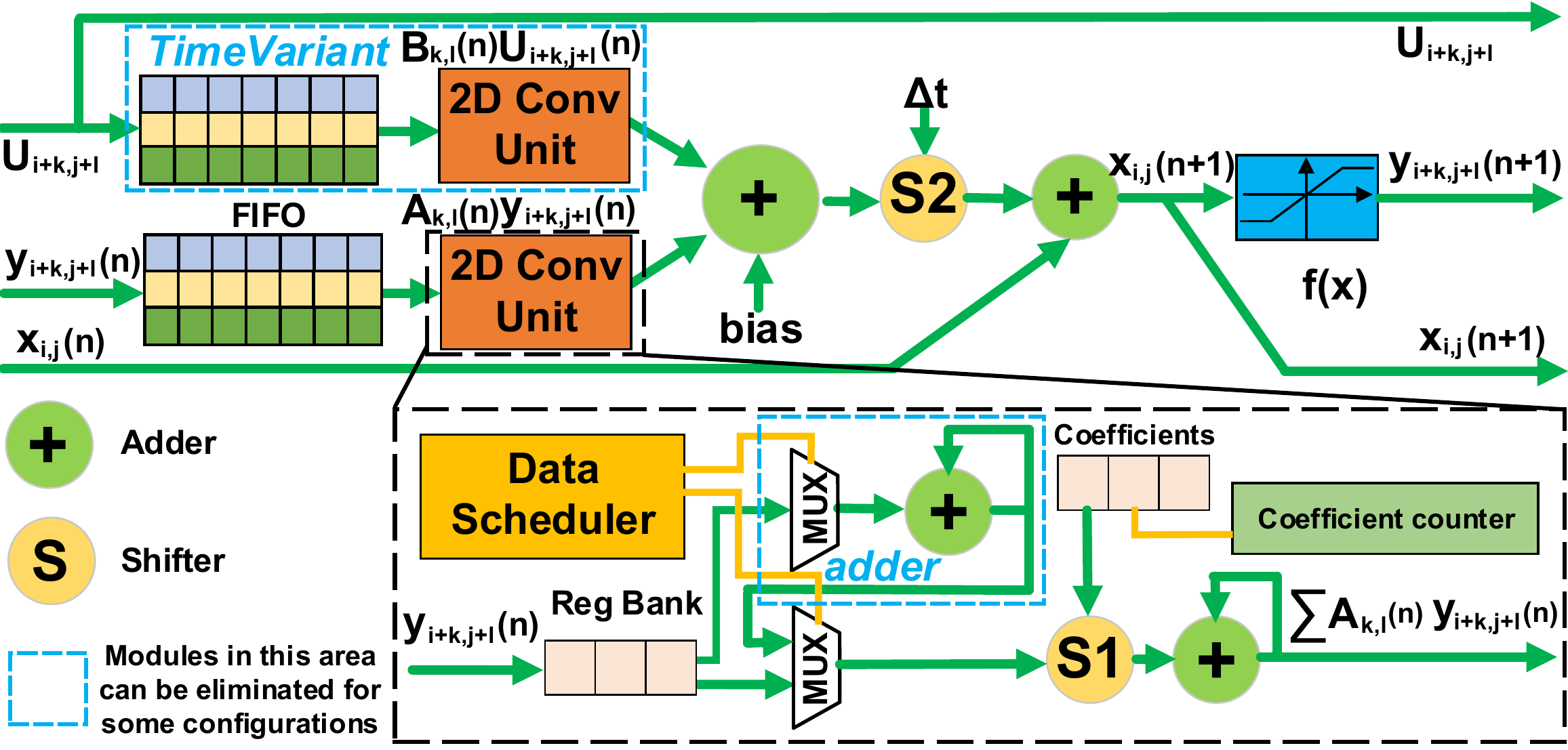}}
\vspace{-1pt}
\caption{Architecture of the optimized stage design.}
\label{fig:PEarchitecture}
\end{center}
\vspace{-22pt}
\end{figure}	

\begin{table}
\vspace{-14pt}
\caption{Comparison of resource utilization between 18-bit multipliers implemented using shifter modules of various configurations $S1(m)$ and $S2(m)$ (with different $m$ as defined in Equation \ref{formula:quantization}, $k$=-$m$ for $S1$, and $k$=0 for $S2$) and a direct implementation of an 18-bit multiplier using LEs and registers.}
\vspace{-1pt}
\label{table:shiftHard}
\begin{center}
\begin{small}
\begin{sc}
\begin{tabular}{p{42pt}<{\centering} p{20pt}<{\centering} p{20pt}<{\centering} p{20pt}<{\centering} p{20pt}<{\centering} p{20pt}<{\centering} p{20pt}<{\centering} p{20pt}<{\centering} p{35pt}<{\centering}}
 \toprule
 Module&$S$1(0)& $S$1(1)& $S$1(2)& $S$1(3) & $S$1(4)& $S$1(5)&$S$2(7) &$Multiplier$\\
 \midrule
LEs           & 39& 44& 50& 80& 109& 105 &80 &676\\
Registers     & 39& 42& 45& 47&  50& 52  &75 &486 \\
\bottomrule
\end{tabular}
\end{sc}
\end{small}
\end{center}
\vspace{-11pt}
\end{table}

Our efficient hardware implementation focuses on the optimization of the stage design as shown in Figure \ref{fig:PEarchitecture}.
Two optimizations are performed: multiplication simplification and data movement optimization.
First, with incremental quantization,
simplification can be achieved by replacing multiplications with shift operations.
The detailed hardware implementation will be discussed in Section \ref{shifterModule}.
Second, when FPGA resource is extremely limited (e.g. for low-end FPGAs), data movement optimization can be performed utilizing the sparsity and repetition in CeNN templates.
As will be discussed later in Section \ref{dataScheduler}, in many applications CeNN templates naturally involves zero or repeated parameters.
With incremental quantization, more zeros are yielded leading to higher sparsity and the small quantization set introduces a larger number of repetitions.
Data movement optimization can minimize the number of computations needed. The details will be discussed in Section \ref{dataScheduler}.

The optimized stage can be configured for both time-invariant templates and time-variant templates.
Note that the FPGA implementation \cite{yildiz2015architecture} is dedicated to CeNN with time-invariant templates, while \cite{martinez2013efficient} is for time-variant.
The $TimeVariant$ part in Figure \ref{fig:PEarchitecture} is specific for time-variant templates, and can be eliminated in the configuration for time-invariant ones.


\subsubsection{Shifter Module}\label{shifterModule}
In Figure \ref{fig:PEarchitecture}, shifter $S1$ is for multiplications in CeNNs and $S2$ is for discrete approximation involved with $\Delta t$ in Equation \ref{formula:CeNND2}.
Usually $\Delta t$ is very small, and the hardware implementation of $S2$ in this paper is designed to support $\Delta t$=$2^{s}$, where $-7\leq s\leq 0$, $s\in \mathbb{Z}$.
Note that when $\Delta t$ is configured to $2^0$ or 1, the computation is transformed to discrete CeNN \cite{harrer1992discrete}.

Table \ref{table:shiftHard} provides an illustrative comparison of resource utilization between multipliers implemented using shifter modules of various configurations and a direct implementation of multiplier using LEs and registers.
It can be noticed that the shifter module consumes much fewer resources than the general implementation,  such that more multiplications can be placed on FPGAs for higher performance and speed.
It should be pointed out that multiple shifters can be adopted in the 2D convolutional module.

\begin{figure}
\vspace{-12pt}
\centering
\subfigure[]{
\label{fig:templateDis1}
\includegraphics[width=0.23\textwidth]{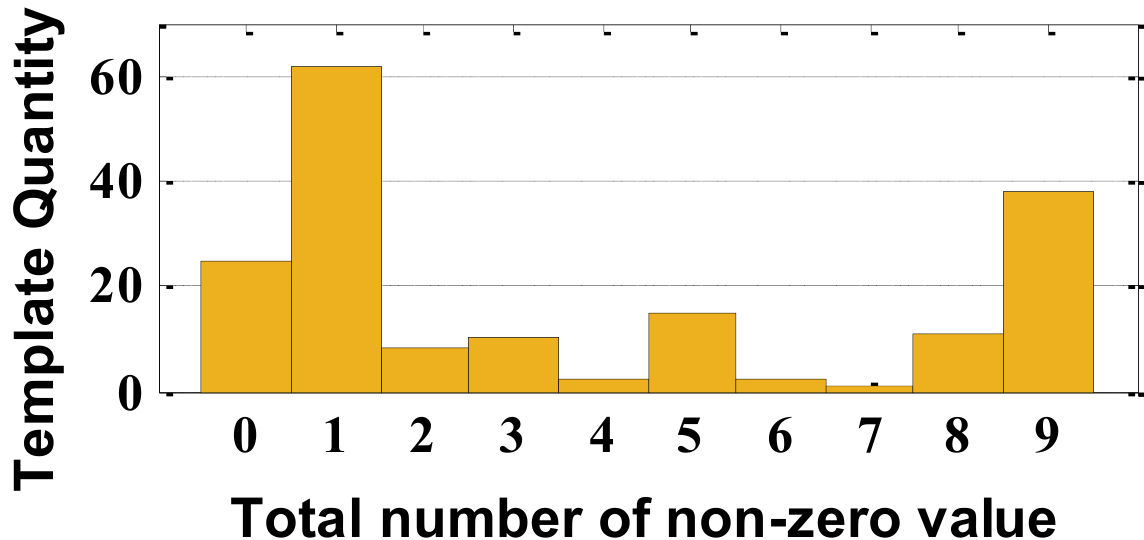}}
\subfigure[]{
\label{fig:templateDis2}
\includegraphics[width=0.23\textwidth]{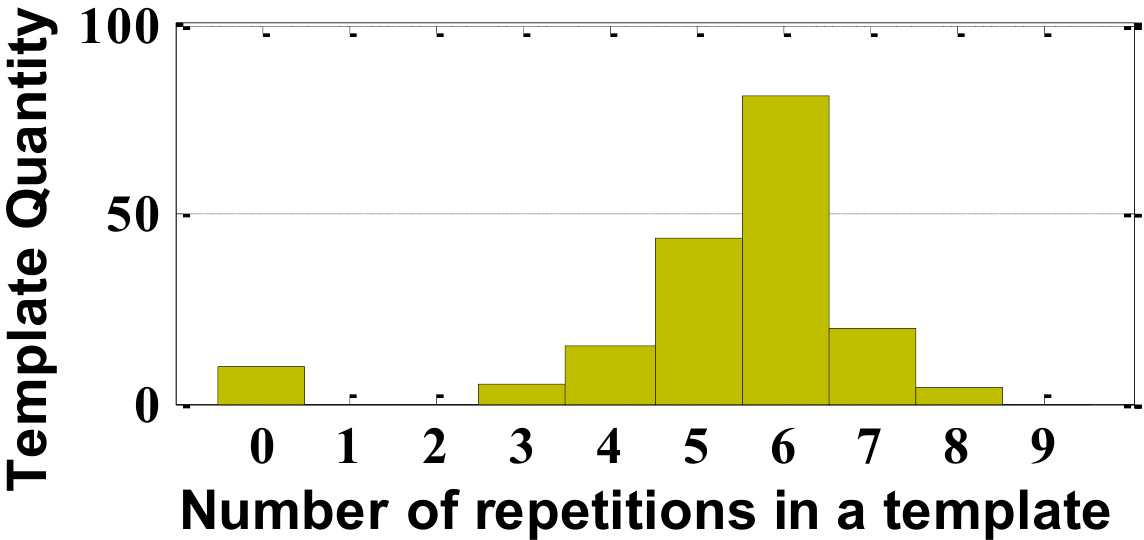}}
\vspace{-11pt}
\caption{Illustration of (a) sparsity and (b) repetition characteristic with 174 CeNN templates.}
\label{fig:templateDis}
\vspace{-5pt}
\end{figure}

\begin{figure}
\vspace{-1pt}
\centering
\centerline{\includegraphics[width=0.6\columnwidth]{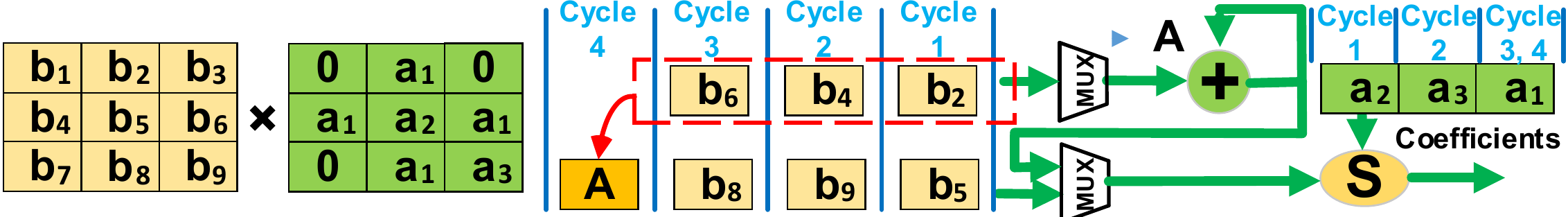}}
\vspace{-1pt}
\caption{Illustration of sparsity-induced and repetition-induced optimizations.}
\label{fig:sparsityAndRepetition}
\vspace{-11pt}
\end{figure}

\subsubsection{Data Scheduler Module}\label{dataScheduler}
Data scheduler module exploits the sparsity and repetition of parameters in CeNN templates.
We analyzed 87 tasks from 79 applications \cite{karacs2010software}, and totally 174 templates are examined (each task has two templates: template $A$ and template $B$).
All the templates are 2D 3$\times$3 each having nine parameters.
The corresponding sparsity and repetition are shown in Figure \ref{fig:templateDis1}.
We can discover that a majority of templates have zero values, and more than half have only three or less non-zero parameters.
Therefore, ignoring multiplications with zeros will give a significant improvement in efficiency.

Figure \ref{fig:templateDis2} depicts the histogram of the parameter repetition in all the 174 templates.
We can see that in most of the templates, about 5-6 parameters are repeated values.
With repeated parameters, we can also take advantage of the associative law for repetition-induced optimization, e.g., $a_1\times b_1+a_1\times b_2+a_1\times b_3=(b_1+b_2+b_3)\times a_1$, and hence three multiplications are optimized to only one.

Note that these optimizations seem to be straightforward and automatic in software synthesis, but
for hardware implementations detailed attention is needed.
An illustration of optimization with sparsity and repetition is shown in Figure \ref{fig:sparsityAndRepetition}.
With sparsity-induced optimization, we only take the non-zero parameters into consideration, and three multiplications can be eliminated.
An adder (only consumes 10 LEs in the design) is utilized to calculate the sum $A$ of $b_2$, $b_4$ and $b_6$ in parallel with the shifter module.
The shifter module calculates $b_5\times a_2$, $b_9\times a_3$, and $b_8\times a_1$ in the first three cycles, and computes $A\times a_1$ in the forth.
Thus, totally it takes four cycles rather than nine cycles to calculate Equation \ref{formula:quantization2}.
Specifically, sparsity-induced optimization reduces the computation time from nine cycles to six, and repetition-induced optimization reduces it from six to four.

The power of sparsity-induced and repetition-induced optimizations varies with different applications.
Note that if the number of shifters adopted in the 2D convolution module is larger than one, repetition-induced optimization can be eliminated as it contributes much less compared with the shifters.
If the number of shifters equals that of the coefficients which is also the situation to achieve the highest throughput,
repetition-induced optimization can also be eliminated as all multiplications can be processed in only one cycle.
Therefore, the two optimizations are only for situations with very limited resources.

\section{Experiments}\label{secExperiment}
In this section, we first evaluate the performance of various incremental quantization strategies discussed in Section \ref{secMethod} with two CPS applications: medical image segmentation for telemedicine and obstacle detection for ADAS.
Then we implement the quantized CeNNs on FPGAs and compare their speed with state-of-the-art works.


\begin{table}
\vskip -0.16in
\caption{Configuration of PSO algorithm.}
\label{table:PSOconfigutation}
\begin{center}
\begin{small}
\begin{sc}
\begin{tabular}{p{20pt}<{\centering} p{10pt}<{\centering} p{10pt}<{\centering} p{10pt}<{\centering} p{30pt}<{\centering} p{20pt}<{\centering} p{20pt}<{\centering} }
\hline
 $N$ & $c_1$ & $c_2$ & $w$ & $iteration$ & $min_d$ & $max_d$ \\
\hline
10    & 1.4& 1.2& 0.8& 500& $-2^m$ &$2^m$\\
\hline
\end{tabular}
\end{sc}
\end{small}
\end{center}
\vspace{-11pt}
\end{table}

\subsection{Performance Evaluation}\label{performanceEvaluation}

\subsubsection{Experimental Setup}\label{Evabinary}
For incremental quantization, a total of 10 incremental quantization strategies are evaluated: five partition strategies (RAN,
PI, WPI, NN (WNN with all weights set to 1), and WNN) in combination with two batch sizes (constant and log-scale).
For compact presentation, we use postfix -C and -L to denote constant and log-scale batch sizes, respectively.
For constant batch size, we set the size to 20\% of the total parameters.
While for log-scale batch size, we set it to half of the remaining un-quantized parameters.
We discuss five quantization set sizes with $m=$0, 1, 2, 3, 4 and $k$ = $-m$ (see Equation \ref{formula:quantization}).

The parameters of PSO algorithm in Equation \ref{formula:PSO} is shown in Table \ref{table:PSOconfigutation}.
The object function designed according to applications will be discussed in the following sections.





\subsubsection{Medical Image Segmentation for Telemedicine}\label{Evadiscussion}

The objective function for medical image segmentation in PSO re-training is shown in Equation \ref{formula:objbinary}, where $output$ and $Ideal$$Output$ are output images of CeNN processing on input images and desired output images, respectively, and $t$ is the number of training pairs, and $area$ is the product of the width and height of the image.
We also adopts the objective function as accuracy to evaluate the quality of segmented images.
The pattern structures of the 3$\times$3 templates $A$ and $B$ are as follows: $A=\{a_0,a_1,a_2; a_3,a_4,a_3;$ $a_2,a_1,a_0\}$, and
$B=\{a_5,a_6,a_7; a_8,a_9,a_8;a_7,a_6,a_5\}$.
The dataset is from the mammographic image analysis society (MIAS) digital mammogram database \cite{suckling1994mammographic}, and two images and its corresponding segmented results are selected as training images as shown in Figure \ref{fig:binaryTraining}, which is the of the same configuration with the work \cite{rouhi2015benign}.
Totally 119 test images are used in the experiment.
Note that as there is no ideal output in the MIAS database, the outputs of the template with double precision are regarded as the ideal outputs.

{
\vspace{-10pt}
\begin{equation}\label{formula:objbinary}
obj=accuracy= \sum_{i=1}^{t} abs(output_i-IdealOutput_i)/area.
\end{equation}
\vspace{-12pt}
}

\begin{figure}
\centering
\centerline{\includegraphics[width=0.5\columnwidth]{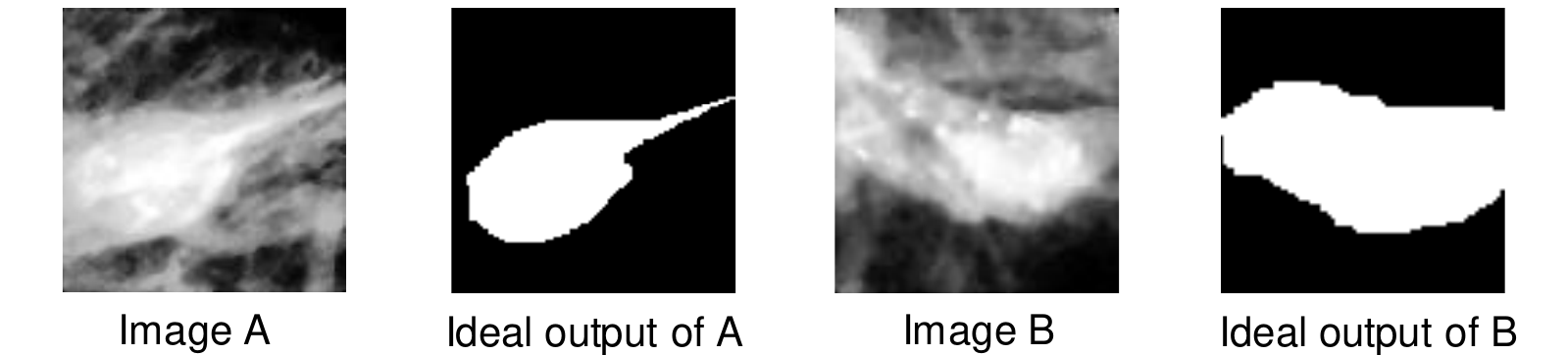}}
\vspace{-1pt}
\caption{Two selected images and their manually segmented images taken from MIAS database to train CeNN.
}
\label{fig:binaryTraining}
\vspace{-1pt}
\end{figure}

\begin{figure}
\centering
\vspace{-16pt}
\subfigure[$m$=2, $k$=-2]{
\label{fig:binaryStraQuan11}
\includegraphics[width=0.3\textwidth]{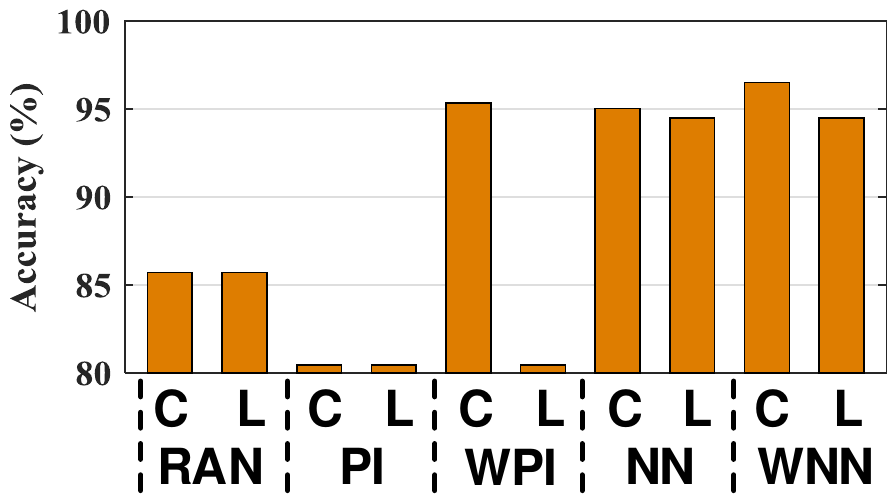}}
\subfigure[NN-L, $k$=-$m$]{
\label{fig:binaryStraQuan12}
\includegraphics[width=0.16\textwidth]{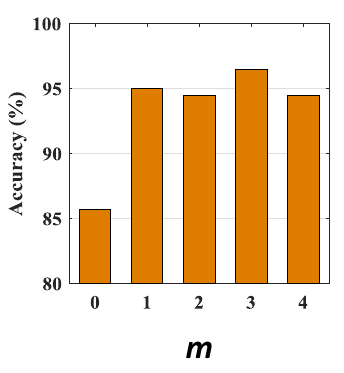}}
\vspace{-11pt}
\caption{Performance comparison between templates with various (a) strategies and (b) quantization sizes $m$ for {biomedical image segmentation}.}
\label{fig:binaryStraQuan1}
\vspace{-5pt}
\end{figure}


We fix the quantization size using $m$ = 2 and $k$ = $-m$, and evaluate all 10 incremental quantization frameworks.
The results are shown in Figure \ref{fig:binaryStraQuan11}.
We can notice that the quantized templates achieve similar accuracy compared with the original template without quantization.
The lowest accuracy is about 12\% lower than that with the original templates.
The highest accuracy is achieved with WNN-C strategy, which is only 3\% lower than that of the original templates. 
{\color{black}{Note that generally PI strategy achieves the best performance for CNNs \cite{zhou2017Incremental}.
However, WNN strategy obtains the best performance for CeNN, and NN strategy also obtains a comparable performance.}}
We can also find that NN and WNN strategy are much stable than PI as NN and WNN can achieve almost the same accuracy for constant and logscale  batch sizes while PI not.
Even random strategy can have a better accuracy than PI in some configurations.
In terms of batch size, constant seems to perform better than log-scale in most cases.
It can be interesting in the future to study this in more detail and figure out a systematic way to decide the optimal strategy.

The impact of batch sizes is presented in Figure \ref{fig:binaryStraQuan12} with the optimal partition WNN-C.
The quantization set size has an interesting relationship with the performance.
First, even when the quantization set is only of three values (-1, 0, 1), the quantized template can still achieve high accuracy.
Second, there exists an optimal $m$ which gives the best performance and $m$=3 for medical image segmentation.
Further increasing $m$ will not provide any performance gain.

%

\subsubsection{Obstacle Detection for ADAS}\label{Evabinary}

We adopts the same objective function as medical image segmentation.
The pattern structures of the 3$\times$3 templates $A$ and $B$ are as follows: $A=\{a_0,a_0,a_0; a_0,a_1,a_0;$ $a_0,a_0,a_0\}$, and
$B=\{a_2,a_2,a_2; a_2,a_3,a_2;a_2,a_2,a_2\}$.
The training dataset is from \cite{feiden2002obstacle} as shown in Figure \ref{fig:Training2}, which is the of the same configuration with the work \cite{rouhi2015benign}.
For test dataset, totally 40 test images are selected from Hlevkin test images collection \cite{Hlevkin}.


\begin{figure}
\centering
\centerline{\includegraphics[width=0.5\columnwidth]{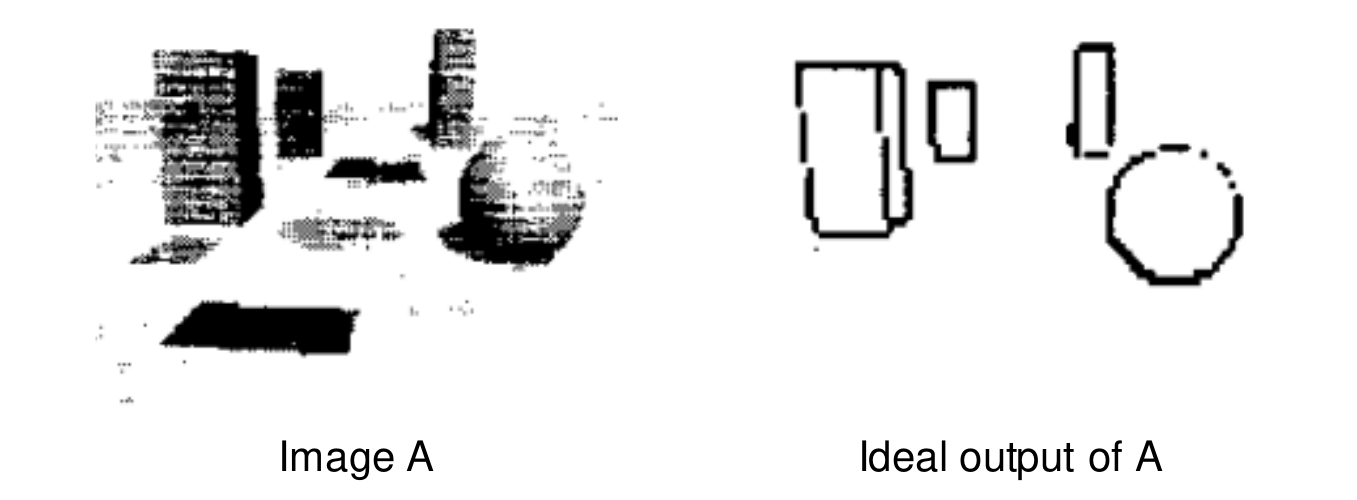}}
\vspace{-14pt}
\caption{Training image and the manually detected image taken from \cite{feiden2002obstacle}.
}
\label{fig:Training2}
\vspace{-12pt}
\end{figure}

\begin{figure}
\centering
\vspace{-1pt}
\subfigure[$m$=2, $k$=-2]{
\label{fig:binaryStraQuan21}
\includegraphics[width=0.28\textwidth]{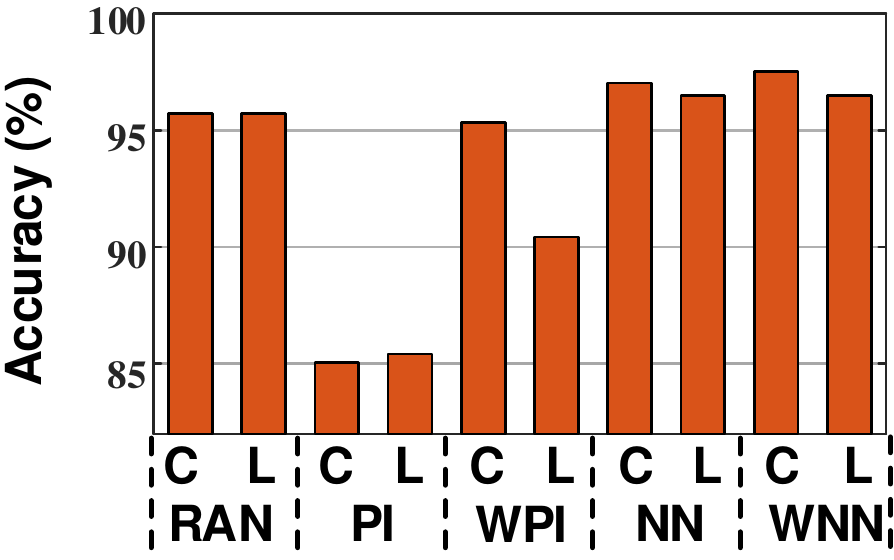}}
\subfigure[NN-L, $k$=-$m$]{
\label{fig:binaryStraQuan22}
\includegraphics[width=0.16\textwidth]{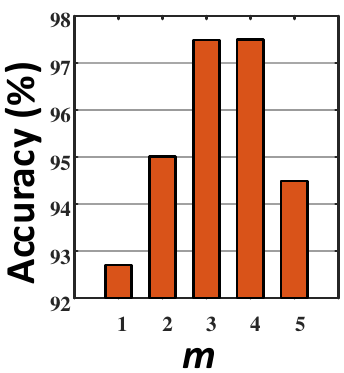}}
\vspace{-6pt}
\caption{Performance comparison between templates with various (a) strategies and (b) quantization sizes $m$ for obstacle detection.}
\label{fig:binaryStraQuan2}
\vspace{-16pt}
\end{figure}


We fix the quantization size using $m$ = 2 and $k$ = $-m$, and evaluate all 10 incremental quantization frameworks.
The results are depicted in Figure \ref{fig:binaryStraQuan21}.
From the figure we can observe that the quantized templates achieve similar accuracy compared with the original template without quantization.
The lowest accuracy is about 15\% lower than that with the original templates.
Like medical image segmentation, the highest accuracy for obstacle detection is achieved with WNN-C strategy, which is only 3\% lower than that of the original templates, 
and constant strategy also performs better than log-scale in most cases.
The impact of batch sizes is presented in Figure \ref{fig:binaryStraQuan22} with the optimal partition WNN-C.
The quantization set size has the same relationship with the performance as that for medical image segmentation.

\subsection{Speed Evaluation Using FPGAs}
In previous section we have evaluated the performance of our CeNN quantization framework in terms of accuracy.
In this section we will evaluate its speed when implemented in FPGAs.
For a fair comparison with existing works \cite{martinez2013efficient}\cite{yildiz2015architecture}\cite{yildiz2016way11111}, we adopt the same configurations of stages and try to place the maximum possible number of stages utilizing our quantized templates.
Note that all the three works share the same architecture for CeNN computation.
The performance of the implementation is evaluated by equivalent computing capacity which is the product of number of stages and the computing capacity of each stage.
The proposed efficient hardware implementation is implemented on an XC4LX25 FPGA.
The data width of the input, state, and output ($u$, $x$, and $y$) is configured to be 18 bits.
The widely-used template size 3$\times$3 is adopted.
Note that general CeNN is adopted for the FPGA implementation, and delayed CeNN is not considered here.
Time-variant templates are configured.
In the implementation, multiplication is achieved with embedded multipliers (more specifically, DSP48 modules on XC4LX25 FPGAs) at first, and shifters are used when there are no more available embedded multipliers.
Considering the routability of FPGAs, the utilization rate of LEs and registers are constrained to be no higher than 80\%.
Note that since different quantization frameworks only affects the performance and do not show significant difference in hardware resource utilization, in this part of experiments we simply use WNN-L with m=5 and k=-5, and other frameworks
should yield almost identical speed.

Three configurations of 2D convolution are discussed: one, three and nine multipliers.
In Table \ref{table:HardwareImple1}, applying our quantization framework can lead to a 1.2x speedup with increased use of LEs (by 17\%) and registers (by 8\%) This allows an additional 4 stages to be placed, with a speedup of 1.2x.

Further taking sparsity-induced optimization into consideration, a speedup of 1.8x is achieved in the 2D convolution module with computations involving with template $A$ for segmentation.
However, no sparsity exists in template B, and there is no overall speedup, as sparsity-induced optimization can only yield speedup when sparsity exists in both templates A and B.
Therefore, the speedup still remain about the same. Yet after the introduction of repetition-induced optimization, the speedup can be further increased to 1.4x with slightly reduced resource usage (due to the reduction of computations needed).
Note that these conclusions are application-specific.
Similar conclusions reside with obstacle detection.
The proposed architecture achieves a little lower clock frequency due to the high resource utilization making placement and routing relatively more difficult.

\begin{table}
\caption{Speed and resource utilization comparisons of the state-of-the-art work \cite{yildiz2016way11111} and ours with {\bf{one multiplier (Mult.)/shifter (Shif.)}} in 2D convolution module, with sparsity-induced optimization and repetition-induced optimization. The numbers in the brackets are the resource utilization rate.}
\vspace{-5pt}
\label{table:HardwareImple1}
\begin{center}
\begin{scriptsize}
\begin{sc}
\begin{tabular}{p{60pt}<{\centering} p{18pt}<{\centering} p{18pt}<{\centering} p{23pt}<{\centering} p{22pt}<{\centering}  }
\hline
 Implementation
 & \multicolumn{1}{c}{\begin{tabular}[c]{@{}c@{}}State-of\\-the-art \\ (1 Mult.)\end{tabular}}
 & \multicolumn{1}{c}{\begin{tabular}[c]{@{}c@{}}Ours \\ (1 shif.)\end{tabular}}
&\multicolumn{1}{c}{\begin{tabular}[c]{@{}c@{}}Ours\\ (1 shif.+\\ sparsity)\end{tabular}}
&\multicolumn{1}{c}{\begin{tabular}[c]{@{}c@{}}Ours\\ (1 shif.+\\ repetition)\end{tabular}}\\
\hline
$\#$ of stages & 24  & 28 & 28   &24  \\
LEs ($\times 10^3$)      & 14.6(60\%)& 18.7(77\%)& 18.7(77\%) &18.4(76\%) \\
Register($\times 10^3$)  & 8.8(40\%)&10.5(48\%)&10.5(48\%)& 9.9(46\%) \\
Embedded Mult.       & 48(100\%)  &48(100\%)&48(100\%) & 48(100\%)   \\
Clock F. (MHz)    & 353&331&331  & 322  \\
Cycles per pixel & 11& 11&11& 8 \\
Speedup     & 1  &\bf{1.2x}& \bf{1.2x}& \bf{1.4x}  \\
\hline
\end{tabular}
\end{sc}
\end{scriptsize}
\end{center}
\vspace{-20pt}
\end{table}

\begin{table}
\caption{Speed and resource utilization comparisons of the state-of-the-art work \cite{yildiz2016way11111} and ours with {\bf{three and nine multipliers(Mult.)/shifter (Shif.)}} in 2D convolution module. The numbers in the brackets are the resource utilization rate.}
\label{table:HardwareImple2}
\begin{center}
\begin{scriptsize}
\begin{sc}
\begin{tabular}{p{60pt}<{\centering} p{28pt}<{\centering} p{28pt}<{\centering} p{28pt}<{\centering} p{28pt}<{\centering} }
\hline
 Implementation&   \multicolumn{1}{c}{\begin{tabular}[c]{@{}c@{}}State-of\\-the-art \\ (3 Mult.)\end{tabular}}  & \multicolumn{1}{c}{\begin{tabular}[c]{@{}c@{}}Ours \\ (3 Shif.)\end{tabular}}  &  \multicolumn{1}{c}{\begin{tabular}[c]{@{}c@{}}State-of\\-the-art \\ (9 Mult.)\end{tabular}}& \multicolumn{1}{c}{\begin{tabular}[c]{@{}c@{}}Ours \\  (9 Shif.)\end{tabular}}\\
\hline
$\#$ of stages                & 6   & 16   &2  &7 \\
LEs($\times 10^3$)        & 3.8(15\%)& 19.6(80\%) &1.4(5\%)&18.2(76\%)\\
Registers($\times 10^3$)    & 2.1(10\%)& 6.5(30\%) &0.6(2\%)&3.6(17\%) \\
Embedded Mult.                         & 48(100\%)  & 48(100\%)   &46(95\%) &48(100\%)\\
Clock F.(MHz) & 337 &320   &361 &343\\
Cycles per pixel & 5& 5 & 1 &1\\
Speedup        & 1 & \bf{2.6x} &1  &\bf{3.5x}\\
\hline
\end{tabular}
\end{sc}
\end{scriptsize}
\end{center}
\vspace{-1pt}
\end{table}

For the configuration of 2D convolution with multiple multipliers,
sparsity-induced and repetition-induced optimizations doing very limited optimizations with multiple multipliers are not involved.
As shown in Table \ref{table:HardwareImple2}, the the state-of-the-art work \cite{yildiz2016way11111} has a very low resource utilization (2\%-15\%) with LEs and registers.
With the abundant resources, 10 and 5 more stages can be placed on FPGAs with shifters as a replacement of multipliers for the implementation configured with three and nine multipliers, respectively, resulting in a speedup of 2.6x and 3.5x.

As the CeNN architecture composed with stage modules are highly extensible, we make a reasonable projections to high-end FPGAs to see how the resources available in an FPGA affect the speedup.
According to existing implementations on FPGAs and resource constraint of 80\% LE and register utilization rate bound,
the clock frequencies are assumed to be the same in the comparison.
The configuration of 2D convolution with nine multipliers is adopted, which has the highest performance.
We select four high-end FPGAs from Altera and Xilinx with about 500,000 to 1,000,000 LEs.
As shown in Table \ref{table:HardwareImple3}, our implementations can achieve a speedup of 1.7x-7.8x.
{\color{black}{Note that the resource consumption of LEs and registers are almost the same for all the implementations, and the speedup varies with the number of embedded multipliers, or more specifically, the ratio of LEs to embedded multipliers.
A high ratio of LEs to embedded multipliers means more LEs can be used to implement shifters resulting with a high speedup.
The highest speedup of 7.8x is due to the fact that the Stratix V E FPGA has the highest rate of LEs to embedded multipliers.}}

\begin{table}
\vspace{-12pt}
\caption{Speed and resource utilization projections to high-end FPGAs of the state-of-the-art work \cite{yildiz2016way11111} and ours with {\bf{nine multipliers/shifters}} in 2D convolution module. The numbers in the brackets are the resource utilization rate.}
\label{table:HardwareImple3}
\begin{center}
\begin{scriptsize}
\begin{sc}
\begin{tabular}{p{60pt}<{\centering} p{28pt}<{\centering} p{28pt}<{\centering} p{28pt}<{\centering} p{28pt}<{\centering} }
\hline
 Implementation&  VC7VX-980T& VC7VX-585T  & Stratix V E &Stratix V GS\\
\hline
$\#$ of stages &352  & 179   &233  &291 \\
LEs($\times 10^3$)        &780(80\%) & 465(80\%)  &718(80\%) &524(80\%)\\
Registers($\times 10^3$)     &170(17\%) & 93(16\%)   &133(15\%) &128(19\%) \\
Embedded Mult.               &3600(100\%) & 1260(100\%)  &704(100\%)  &3926(100\%)\\
Speedup        &\bf{2.3x} & \bf{3.3x}  &\bf{7.8x} &\bf{1.7x}\\
\hline
\end{tabular}
\end{sc}
\end{scriptsize}
\end{center}
\vspace{-20pt}
\end{table}

\section{Conclusions}\label{secConclusion}
In this paper, we propose CeNN quantization for computation reduction for CPS, particularly telemedicine and ADAS.
The powers-of-two based incremental quantization adopts an iterative procedure including parameter partition, parameter quantization, and re-training to produce templates with values being powers of two.
We propose a few quantization strategies based on the unique CeNN computation patterns.
Thus, multiplications are transformed to shift operations, which are much more resource-efficient than general embedded multipliers.
{\color{black}{
Furthermore, based on CeNN template structures, sparsity-induced and repetition-induced optimizations for quantized templates are also exploited for situations where resources are extremely limited.
}}
Experimental results on medical image segmentation for telemedicien and obstacle detection for ADAS show that the proposed quantization framework can achieve similar performance compared with that using original templates without optimization, and the implementation with incremental quantization can achieve a speedup up to 7.8x compared with the state-of-the-art FPGA implementations.
{\color{black}{We also discover that unlike CNNs, the optimal strategy of CeNNs is weighted nearest neighbor strategy other than pruning-inspired strategy.}}

\bibliographystyle{plain}

%% file: letter.bbl
\begin{thebibliography}{}

\end{thebibliography}


\begin{thebibliography}{10}

\bibitem{cpsSurvey}
Khaitan et al., Design Techniques and Applications of Cyber Physical Systems: A Survey, IEEE Systems Journal, 2014.

\bibitem{carey2013mixed}
S.~J. Carey, D.~R. Barr, B.~Wang, A.~Lopich, and P.~Dudek.
\newblock Mixed signal simd processor array vision chip for real-time image
  processing.
\newblock {\em Analog Integrated Circuits and Signal Processing},
  77(3):385--399, 2013.

\bibitem{chae2014medical}
S.-H. Chae, D.~Moon, D.~G. Lee, and S.~B. Pan.
\newblock Medical image segmentation for mobile electronic patient charts using
  numerical modeling of iot.
\newblock {\em Journal of Applied Mathematics}, 2014, 2014.

\bibitem{xu2018mda}
X.~Xu, F.~Lin, and et~al., ``Mda: A reconfigurable memristor-based distance
  accelerator for time series mining on data centers,'' \emph{TCAD}, 2018.

\bibitem{xu2018scaling}
X.~Xu, Y.~Ding, and et~al., ``Scaling for edge inference of deep neural
  networks,'' \emph{Nature Electronics}, vol.~1, no.~4, p. 216, 2018.

\bibitem{xu2018quantization}
X.~Xu, Q.~Lu, and et~al., ``Quantization of fully convolutional networks for
  accurate biomedical image segmentation,'' \emph{CVPR}, 2018.

\bibitem{chen2006image}
H.-C. Chen, Y.-C. Hung, C.-K. Chen, T.-L. Liao, and C.-K. Chen.
\newblock Image-processing algorithms realized by discrete-time cellular neural
  networks and their circuit implementations.
\newblock {\em Chaos, Solitons \& Fractals}, 29(5):1100--1108, 2006.

\bibitem{chua2002cellular}
L.~O. Chua and T.~Roska.
\newblock {\em Cellular neural networks and visual computing: foundations and
  applications}.
\newblock Cambridge university press, 2002.

\bibitem{courbariaux2016binarized}
M.~Courbariaux, I.~Hubara, D.~Soudry, R.~El-Yaniv, and Y.~Bengio.
\newblock Binarized neural networks: Training deep neural networks with weights
  and activations constrained to+ 1 or-1.
\newblock {\em arXiv preprint arXiv:1602.02830}, 2016.

\bibitem{dohler2008cellular}
F.~Dohler, F.~Mormann, B.~Weber, C.~E. Elger, and K.~Lehnertz.
\newblock A cellular neural network based method for classification of magnetic
  resonance images: towards an automated detection of hippocampal sclerosis.
\newblock {\em Journal of neuroscience methods}, 170(2):324--331, 2008.

\bibitem{duraisamy2014cellular}
M.~Duraisamy and F.~M.~M. Jane.
\newblock Cellular neural network based medical image segmentation using
  artificial bee colony algorithm.
\newblock In {\em Green Computing Communication and Electrical Engineering
  (ICGCCEE), 2014 International Conference on}, pages 1--6. IEEE, 2014.

\bibitem{harrer1992discrete}
H.~Harrer and J.~A. Nossek.
\newblock Discrete-time cellular neural networks.
\newblock {\em International Journal of Circuit Theory and Applications},
  20(5):453--467, 1992.

\bibitem{harrer1994current}
H.~Harrer, J.~A. Nossek, T.~Roska, and L.~O. Chua.
\newblock A current-mode dtcnn universal chip.
\newblock In {\em Circuits and Systems, 1994. ISCAS'94., 1994 IEEE
  International Symposium on}, volume~4, pages 135--138. IEEE, 1994.

\bibitem{hubara2016binarized}
I.~Hubara, M.~Courbariaux, D.~Soudry, R.~El-Yaniv, and Y.~Bengio.
\newblock Binarized neural networks.
\newblock In {\em Advances in Neural Information Processing Systems}, pages
  4107--4115, 2016.

\bibitem{karacs2010software}
K.~Karacs, G.~Cserey, Zarndy, P.~Szolgay, C.~Rekeczky, L.~Kek, V.~Szab,
  G.~Pazienza, and T.~Roska.
\newblock Software library for cellular wave computing engines.
\newblock {\em Cellular Sensory and Wave Computing Laboratory of the Computer
  and Automation Research Institute}, 2010.

\bibitem{lee201124}
S.~Lee, M.~Kim, K.~Kim, J.-Y. Kim, and H.-J. Yoo.
\newblock 24-gops 4.5-$mm^2$ digital cellular neural network for rapid visual
  attention in an object-recognition soc.
\newblock {\em IEEE transactions on neural networks}, 22(1):64--73, 2011.

\bibitem{li2011edge}
H.~Li, X.~Liao, C.~Li, H.~Huang, and C.~Li.
\newblock Edge detection of noisy images based on cellular neural networks.
\newblock {\em Communications in Nonlinear Science and Numerical Simulation},
  16(9):3746--3759, 2011.

\bibitem{magdalena2015use}
M.~Magdalena and U.~G.~B. Bujnowska-Fedak.
\newblock Use of telemedicine-based care for the aging and elderly: promises
  and pitfalls.
\newblock {\em Smart homecare Technology \& telehealth}, 3:91--105, 2015.

\bibitem{manatunga2015sp}
D.~Manatunga, H.~Kim, and S.~Mukhopadhyay.
\newblock Sp-cnn: A scalable and programmable cnn-based accelerator.
\newblock {\em IEEE Micro}, 35(5):42--50, 2015.

\bibitem{manganaro2012cellular}
G.~Manganaro, P.~Arena, and L.~Fortuna.
\newblock {\em Cellular neural networks: chaos, complexity and VLSI
  processing}, volume~1.
\newblock Springer Science \& Business Media, 2012.

\bibitem{martinez2013efficient}
J.~J. Martnez, J.~Garrigs, J.~Toledo, and J.~M. Ferrndez.
\newblock An efficient and expandable hardware implementation of multilayer
  cellular neural networks.
\newblock {\em Neurocomputing}, 114:54--62, 2013.

\bibitem{muller2016improved}
J.~Muller, R.~Wittig, J.~Muller, and R.~Tetzlaff.
\newblock An improved cellular nonlinear network architecture for binary and
  greyscale image processing.
\newblock {\em IEEE Transactions on Circuits and Systems II: Express Briefs},
  2016.

\bibitem{porter2007reconfigurable}
R.~Porter, J.~Frigo, A.~Conti, N.~Harvey, G.~Kenyon, and M.~Gokhale.
\newblock A reconfigurable computing framework for multi-scale cellular image
  processing.
\newblock {\em Microprocessors and Microsystems}, 31(8):546--563, 2007.

\bibitem{potluri2011cnn}
S.~Potluri, A.~Fasih, L.~K. Vutukuru, F.~Al~Machot, and K.~Kyamakya.
\newblock Cnn based high performance computing for real time image processing
  on gpu.
\newblock In {\em Nonlinear Dynamics and Synchronization (INDS) \& 16th Int'l
  Symposium on Theoretical Electrical Engineering (ISTET), 2011 Joint 3rd Int'l
  Workshop on}, pages 1--7. IEEE, 2011.

\bibitem{rastegari2016xnor}
M.~Rastegari, V.~Ordonez, J.~Redmon, and A.~Farhadi.
\newblock Xnor-net: Imagenet classification using binary convolutional neural
  networks.
\newblock In {\em European Conference on Computer Vision}, pages 525--542.
  Springer, 2016.

\bibitem{xu2017segmentation}
X.~Xu, Q.~Lu, T.~Wang, J.~Liu, C.~Zhuo, S.~Hu, and Y.~Shi.
\newblock Empowering Mobile Telemedicine with Compressed Cellular Neural Networks.
\newblock In {\em Proc. of IEEE/ACM 2017 International Conference On Computer-Aided Design}.
  IEEE/ACM, 2017.

\bibitem{rodriguez2004ace16k}
A.~Rodrguez-Vzquez, G.~Lin-Cembrano, L.~Carranza, E.~Roca-Moreno,
  R.~Carmona-Galn, F.~Jimnez-Garrido, R.~Domnguez-Castro, and S.~E. Meana.
\newblock Ace16k: the third generation of mixed-signal simd-cnn ace chips
  toward vsocs.
\newblock {\em IEEE Transactions on Circuits and Systems I: Regular Papers},
  51(5):851--863, 2004.

\bibitem{rouhi2015benign}
R.~Rouhi, M.~Jafari, S.~Kasaei, and P.~Keshavarzian.
\newblock Benign and malignant breast tumors classification based on region
  growing and cnn segmentation.
\newblock {\em Expert Systems with Applications}, 42(3):990--1002, 2015.

\bibitem{han2016deep}
H.~Song, P.~Jeff, T.~John, and W.~J. Dally.
\newblock Deep compression: Compressing deep neural networks with pruning,
  trained quantization and huffman coding.
\newblock In {\em 4th International Conference on Learning Representations},
  2016.

\bibitem{suckling1994mammographic}
J.~Suckling, J.~Parker, D.~Dance, S.~Astley, I.~Hutt, C.~Boggis, I.~Ricketts,
  E.~Stamatakis, N.~Cerneaz, S.~Kok, et~al.
\newblock The mammographic image analysis society digital mammogram database.
\newblock In {\em Exerpta Medica. International Congress Series}, volume 1069,
  pages 375--378, 1994.

\bibitem{wong2011comparing}
H.~Wong, V.~Betz, and J.~Rose.
\newblock Comparing fpga vs. custom cmos and the impact on processor
  microarchitecture.
\newblock In {\em Proceedings of the 19th ACM/SIGDA international symposium on
  Field programmable gate arrays}, pages 5--14. ACM, 2011.

\bibitem{wu2016quantized}
J.~Wu, C.~Leng, Y.~Wang, Q.~Hu, and J.~Cheng.
\newblock Quantized convolutional neural networks for mobile devices.
\newblock In {\em Proceedings of the IEEE Conference on Computer Vision and
  Pattern Recognition}, pages 4820--4828, 2016.

\bibitem{yildiz2015architecture}
N.~Yildiz, E.~Cesur, K.~Kayaer, V.~Tavsanoglu, and M.~Alpay.
\newblock Architecture of a fully pipelined real-time cellular neural network
  emulator.
\newblock {\em IEEE Transactions on Circuits and Systems I: Regular Papers},
  62(1):130--138, 2015.

\bibitem{liu2018efficient}
Z.~Liu, C.~Zhuo, and X.~Xu, ``Efficient segmentation method using quantised and
  non-linear cenn for breast tumour classification,'' \emph{Electronics
  Letters}, 2018.

\bibitem{xu2018efficient}
X.~Xu, Q.~Lu, and et~al., ``Efficient hardware implementation of cellular
  neural networks with incremental quantization and early exit,'' \emph{JETC},
  vol.~14, no.~4, p.~48, 2018.

\bibitem{xu2018accelerating}
X.~Xu, F.~Lin, and et~al., ``Accelerating dynamic time warping with
  memristor-based customized fabrics,'' \emph{TCAD}, vol.~37, no.~4, pp.
  729--741, 2018.
  
\bibitem{yildiz2016way11111}
N.~Yildiz, E.~Cesur, and V.~Tavsanoglu.
\newblock On the way to a third generation real-time cellular neural network
  processor.
\newblock {\em CNNA 2016}, 2016.

\bibitem{zhou2017Incremental}
A.~Zhou, A.~Yao, Y.~Guo, L.~Xu, and Y.~Chen.
\newblock Incremental network quantization: Towards lossless cnns with
  low-precision weights.
\newblock In {\em 5th International Conference on Learning Representations},
  2017.

\bibitem{xu2018dac}
X.~Xu, X.~Zhang, and et~al., ``Dac-sdc low power object detection challenge for
  uav applications,'' \emph{arXiv preprint arXiv:1809.00110}, 2018.

\bibitem{feiden2002obstacle}
D.~Feiden and R.~Tetzlaff.
\newblock Obstacle detection in planar worlds using cellular neural networks.
\newblock In {\em Cellular Neural Networks and Their Applications, 2002.(CNNA
  2002). Proceedings of the 2002 7th IEEE International Workshop on}, pages
  383--390. IEEE, 2002.

\bibitem{Hlevkin}
Hlevkin.
\newblock http://www.hlevkin.com/06testimages.htm, 2017.


\end{thebibliography}
